\pdfoutput=1

\documentclass[10pt,twocolumn,letterpaper]{article}

\usepackage{iccv}
\usepackage{times}
\usepackage{epsfig}
\usepackage{graphicx}
\usepackage{amsmath}
\usepackage{amssymb}
\usepackage{xfrac}
\usepackage{algorithm}
\usepackage{algpseudocode}
\usepackage[normalem]{ulem}
\usepackage[breaklinks=true,colorlinks,bookmarks=false]{hyperref}

\newtheorem{proposition}{Proposition}
\newtheorem{corollary}{Corollary}

\newcommand{\norm}[1]{\left\Vert#1\right\Vert}

\newcommand{\set}[1]{\left\{#1\right\}}
\newcommand{\parr}[1]{\left (#1\right )}
\newcommand{\brac}[1]{\left [#1\right ]}

\newcommand{\myspan}[1]{\mathrm{span}\set{ #1 } }
\newcommand{\Real}{\mathbb R}

\newcommand{\eps}{\varepsilon}

\newcommand{\too}{\rightarrow}

\newcommand{\wt}[1]{\widetilde{#1}} 

\def \bdk{\mathcal{D}_{\mu}}
\def \T{\mathcal{T}}
\def \V{\mathcal{V}}
\def \E{\mathcal{E}}
\def \F{\mathcal{F}}
\def \vecx{\mathbf{x}}
\def \vecv{\mathbf{v}}
\def \vect{\mathbf{t}}
\def \vece{\mathbf{e}}
\def \vecp{\mathbf{p}}
\def \vecq{\mathbf{q}}
\def \vech{\mathbf{h}}
\def \veczero{\mathbf{0}}

\iccvfinalcopy 

\def\iccvPaperID{1697} 

\ificcvfinal\fi

\begin{document}

\title{Wide baseline stereo matching with convex bounded-distortion constraints}


\author{Meirav Galun\thanks{Equal contributors}\ \ \thanks{The Weizmann Institute of Science, Israel} \and Tal Amir\footnotemark[1]\ \ \footnotemark[2] \and Tal Hassner\thanks{The Open University, Israel}  \and Ronen Basri\footnotemark[2] \and Yaron Lipman\footnotemark[2]}

\maketitle

\begin{abstract}
Finding correspondences in wide baseline setups is a challenging problem. Existing approaches have focused largely on developing better feature descriptors for correspondence and on accurate recovery of epipolar line constraints. This paper focuses on the challenging problem of finding correspondences once approximate epipolar constraints are given.
We introduce a novel method that integrates a deformation model. Specifically, we formulate the problem
as finding the largest number of corresponding points related by a bounded distortion map
that obeys the given epipolar constraints. We show that, while the set of bounded distortion maps is not convex, the subset of maps that obey the epipolar line constraints is convex, allowing us to introduce an efficient algorithm for matching. We further utilize a robust cost function for matching and employ majorization-minimization for its optimization. Our experiments indicate that our method finds significantly more accurate maps than existing approaches.

\end{abstract}

\section{Introduction}

Finding point correspondences in image pairs of a static scene is a classical problem in stereo and structure from motion (SFM). Finding correspondences in wide baseline setups, i.e., when the cameras' focal centers are distant, is particularly challenging. Images obtained in such setups are generally subject to significant distortion and their content may differ substantially also due to occlusion.

The problem of wide baseline stereo matching has received significant attention in recent years (see a brief review in Section~\ref{sec:review}).
Existing approaches have focused largely on developing better feature descriptors for correspondence and on accurate recovery of epipolar line
constraints. However, although challenging, the problem of finding correspondences once the epipolar geometry has been estimated has not yet received sufficient attention. 

In this paper we introduce a novel method for finding correspondences in wide baseline image pairs of a static scene. Noting that matching is often ambiguous even when epipolar constraints are taken into account, we propose to address the problem by using deformation maps to model geometric changes along epipolar lines. Specifically, given two images and an estimated fundamental matrix, our algorithm seeks to compute a geometric map that relates the images and satisfies two requirements; First, it should respect the epipolar constraints, and, secondly, we bound the amount of distortion that the mapping can exert locally. We refer to such a map by  \textit{epipolar consistent bounded-distortion} (EBD) map. Our core theoretical contribution is in showing that, while the set of maps whose distortion is bounded is non-convex, its intersection with maps that satisfy the epipolar constraints (with an ordering assumption \cite{Baker81}) is convex, allowing us to introduce an efficient matching algorithm.

{\em Bounded distortion} (BD) maps are continuous, locally injective transformations whose conformal distortion at every point (defined as the condition number of their Jacobian matrices) is bounded. Intuitively, the conformal distortion measures how different the local map is from a similarity transformation, \ie, how much local aspect ratio is changed. Bounding the conformal distortion is motivated by the following observation. Suppose two cameras are set so that their image planes are parallel (including as special case rectified setups). For any fronto-parallel plane it can be readily verified that its projections  onto the two image planes are related by a similarity transformation. Therefore such projections undergo no distortion. Bounding the distortion in these setups therefore limits the slant and tilt of the recovered planes.

To formulate our solution we define a cost function that seeks an EBD map that maximizes the number of matches. We optimize this robust objective using majorization-minimization. The use of a robust objective allows us to recover when certain portions of the images are distorted beyond the bounds allowed by our algorithm or when the set of initial correspondences include outliers.

We have tested our method on datasets containing pairs of images with ground truth matches and compared it to several state-of-the-art methods. Our method consistently outperformed these methods.

\section{Previous work}
\label{sec:review}

The problem of wide baseline stereo matching has been approached by a number of studies. Considerable effort has been put into designing better features and descriptors and into utilizing them to estimating the fundamental matrix. Several studies have used affine invariant features \cite{Van_Gool_2000,Shah_2003}. A wide variety of alternatives to the SIFT descriptor \cite{SIFT} have been proposed, emphasizing speed (e.g, the Daisy descriptor~\cite{Daisy_2010}) or invariance to extreme transformations such as scale changes~\cite{hassner2012sifts}. Other studies have utilized line segments~\cite{FeaturesLineSegments_2005} and regional features (e.g., MSER~\cite{Matas_2004} and texture-based descriptors~\cite{Schaffalitzky_2001}). \cite{Zisserman_1998} groups coplanar points by identifying homographies and uses them to estimate epipolar lines. A few of those descriptors were designed to also account for occlusion (e.g.,~\cite{Daisy_2010,segmentation-aware_2013}). Finally, a number of studies have approached the problem from a multiview perspective~\cite{Strecha_PDE_2003,Ferrari_2003}.

Relevant to our work also are generic methods for robust, dense matching, based on a variety of point-feature and regional descriptors, such as the SIFT-flow~\cite{liu2008sift,SiftFlow_2011}, patch-match~\cite{PatchMatch_2009}, NRDC~\cite{NRDC_2011}, LDOF~\cite{brox2009large} and, more recently, SPM \cite{kim2013deformable}, as well as models of deformation (e.g.,~\cite{Berg05,Duchenne11,Leordeanu05}), which can potentially be applied in a wide baseline setting. Another recent study \cite{lin2011smoothly} proposed an algorithm for mosaic stitching by finding a map that smoothly departs from a global affine transformation. Our experiments include comparison to \cite{Leordeanu05} and~\cite{SiftFlow_2011} modified to seek matches near corresponding epipolar lines. We show the results of our method are superior to these methods even despite these modifications, suggesting that our global model of deformation provides a more suitable model for wide baseline stereo.


Our model of deformation maps is derived from the work of~\cite{lipman12}, that proposed an approach for optimizing functionals over bounded distortion transformations using sequences of convex optimization. \cite{lipman14} further used this approach for robust feature matching in general pairs of images~(analogous to RANSAC~\cite{Fischler81}, but allowing many degrees of freedom). Our work shows that the set of EBD maps are convex, allowing us to introduce an efficient algorithm that is less sensitive to initialization.

\section{Method}
\label{sec:method}

In this section we describe our algorithmic approach to the problem of wide baseline image matching. We assume we are given two images $I,J\subset\Real^2$, with their fundamental matrix $F$ either supplied as input or computed automatically, e.g., using RANSAC~\cite{Fischler81}. Our goal is to find a map $\Phi$ from $I$ to $J$ that relates corresponding points in the two images; \ie, for every pair of corresponding points, $(\vecp,\vecq) \in I\times J$, the desired map satisfies $\Phi(\vecp)=\vecq$. We start with a large set of candidate corresponding pairs of points $(\vecp_m,\vecq_m)\in I\times J$, $m=1,...,n$. Then, we search for a map $\Phi$, from the family of epipolar $\mu$-bounded distortion mappings $\bdk$ (defined below) that matches as many  pairs $(\vecp_m,\vecq_ m)$ as possible. Specifically, we aim at optimizing
\begin{subequations} \label{e:opt}
\begin{align} \label{e:opt1}
\min_{\Phi} & \,\,\,\sum_{m=1}^n   \norm{\Phi(\vecp_m)-\vecq_m}_2^0 \\ \label{e:opt2}
\mathrm{s.t.} & \,\,\, \Phi\in \bdk,
\end{align}
\end{subequations}
where for $\vecv \in \Real^{2}$ the norm $\norm{\cdot}_2^0$ is defined by: $\norm{\vecv}_2^0=1$ if $\vecv \ne \veczero$, and $\norm{\vecv}_2^0=0$ otherwise. The optimization problem \eqref{e:opt} strives to maximize the number of matched pairs under the deformation model. This can be seen by noting that the energy \eqref{e:opt1} counts how many pairs $(\vecp_m,\vecq_m)$ are not matched by $\Phi$. Similarly to \cite{lipman14}, we solve \eqref{e:opt} by: 1) computing a set of candidate pairs of correspondences $(\vecp_m,\vecq_m)$; and 2) optimizing \eqref{e:opt} using an iterative re-weighted least-squares (IRLS) approach. However, differently from previous work, we devise a novel formulation of the Bounded Distortion deformation model that is shown to be \emph{convex} when matching images under the epipolar constraint. The convex model facilitates the optimization of \eqref{e:opt}, allows considerably faster optimization times, incorporates epipolar constraints, and does not require any particular initialization or convexification. We explain the deformation model next.

\subsection{Convex Epipolar BD Deformations}
At the core of our method is a convex characterization of the space $\bdk$ of epipolar BD deformations. In a nut-shell, $\bdk$ is a one parameter family of non-rigid deformations that allow bounded amount of distortion and respect epipolar constraints. To formulate $\bdk$ we introduce a \emph{triangulation} $\T=(\V,\E,\F)$ on image $I$, where $\V=\set{\vecv_i}\subset I$ is the vertex set, $\E=\set{e_{ij}}$ the edge set, and $\F=\set{f_{ijk}}$ the triangles (faces).

A mapping $\Phi\in\bdk$ is represented by prescribing new locations to the vertices of the triangulation in the second image, $\wt{\V}=\set{\wt{\vecv}_i}\subset J$. The mapping $\Phi$ is defined as the unique piecewise-linear (PL) mapping satisfying $\Phi(\vecv_i)=\wt{\vecv}_i$. We denote by $\Phi_{ijk}\doteq\Phi\vert_{f_{ijk}}$ the affine map of the restriction of $\Phi$ to the triangle $f_{ijk} \in \F$.

Using the entire collection of PL mappings $\{\Phi\}$ defined on a triangulation $\T$ is way too general as every vertex is allowed to move arbitrarily and in the context of stereo this will allow unreasonable geometries to be considered. Instead, we will restrict our attention to a one parameter family of mapping spaces $\bdk$ that translate to a reasonable assumption of the scene's geometry. In particular, in addition to imposing epipolar line constraints, 
we suggest to bound the deviation of the affine maps $\Phi_{ijk}$ from similarity transformations using a parameter $0<\mu<1$ as is defined below. We next derive this constraint for a single affine transformation and later show how to set the constraints for the entire triangulation $T$ to define $\bdk$.

\subsubsection{Epipolar Bounded-Distortion affine map}
We now focus on a single affine map. A general planar affine map can be written uniquely as
\begin{equation}\label{e:affine}
f(\vecx)=B \vecx+C \vecx +\vect
\end{equation}
where,
$$B=\small\begin{pmatrix}a & b \\
-b & a \\
\end{pmatrix}\,, \quad C = \small\begin{pmatrix}c & d \\
d & -c \\
\end{pmatrix}\, , \quad \vect = \begin{pmatrix}t^1 \\
t^2 \\
\end{pmatrix}$$ are a similarity matrix, an anti-similarity matrix (\ie, a reflected similarity), and a translation vector, respectively \cite{lipman12,lipman14}. The ratio of Frobenious norms of the anti-similarity and similarity parts, \ie, $$\mu_f=\frac{\norm{C}}{\norm{B}}=\sqrt{\frac{c^2+d^2}{a^2+b^2}}$$ provides a natural scale-invariant measure for deviation of $f$ from a similarity. In fact, $$K_f=\frac{1+\mu_f}{1-\mu_f}$$ is the \emph{conformal distortion} of the affine map, which equals the ratio of the maximal singular value to the minimal singular value (\ie, the condition number) of the linear part of the affine map, $B+C$.  We therefore set the $\mu$-Bounded Distortion constraint,
\begin{equation}\label{e:bd}
\mu_f \leq \mu
\end{equation}
where as mentioned above $0<\mu<1$ is a parameter of the deformation space. We note that an affine map satisfying \eqref{e:bd} is also orientation preserving since $2^{1/2}\det(B+C)=\norm{B}^2-\norm{C}^2$ and $0<\mu<1$.

The Bounded-Distortion constraint \eqref{e:bd} is not convex and requires some convexification to work with in practice~\cite{lipman12}. However, surprisingly, it becomes convex when we intersect this constraint with the epipolar line constraints (assuming epipolar line pairs can be oriented, as we explain below). More generally, when the affine map $f$ is known to map some directed line $\ell_1$ (\eg, epipolar line) to another directed line $\ell_2$, while preserving the direction, then  Eq.~\eqref{e:bd} can be formulated as a convex constraint in $B,C$, see Figure \ref{fig:epibd} for an illustration. We summarize this in a Proposition,
\begin{proposition}\label{prop:basic}
The collection of $\mu$-Bounded-Distortion planar affine transformations that map a directed line $\ell_1$ to another directed line $\ell_2$  is convex. \end{proposition}
%
We start by proving the proposition for the case that the directed lines both coincide with the $X$-axis with the positive direction, $$\ell_1= \ell_2 = \ell = \myspan{\vece_1}$$ where $\vece_1=(1,0)^T$. By assumption we have in particular that $f(\veczero),f(\vece_1)\in\ell$ and $\vece_1^Tf(\veczero)< \vece_1^Tf(\vece_1)$. This implies that
\begin{equation}\label{e:a_and_b}
\vece_2^T \vect=0\, , \quad d=b\, , \quad a+c>0
\end{equation}
where $\vece_2=(0,1)^T$. Plugging this into \eqref{e:bd}, squaring and rearranging we get
\begin{equation}\label{e:almost_cone}
(1-\mu^2)b^2 + c^{2}\leq \mu^2 a^2.
\end{equation}
If we show that $a>0$ then taking the square-root of both sides of \eqref{e:almost_cone} leads to a (convex) second-order cone (SOC) constraint,
\begin{equation}\label{e:cone}
\sqrt{(1-\mu^2)b^2+c^{2}} \leq \mu a.
\end{equation}
Indeed, since $a+c > 0$ and \eqref{e:almost_cone} implies that $|a|>|c|$ we must have $a>0$. We have therefore shown that any affine map \eqref{e:affine} that satisfies the assumption \eqref{e:bd} and maps the real axis $\ell$ to itself  by preserving the positive direction has to satisfy \eqref{e:a_and_b} and \eqref{e:cone}. In the other direction, any non-zero affine map that satisfy \eqref{e:a_and_b} and  \eqref{e:cone} maps $\ell$ to itself while preserving the positive direction (since $a+c>0$) and satisfies \eqref{e:bd}.

For general directed lines $\ell_1,\ell_2$ we can represent any affine map $f^*$ satisfying the assumptions of Proposition \ref{prop:basic} as
\begin{equation}\label{e:change_of_coords}
f^* =g_2\circ f \circ g_1^{-1}
\end{equation}
where $g_i$, $i=1,2$, are similarities that map the $X$-axis $\ell$ (with positive direction) to $\ell_i$, and $f$ is $\mu$-Bounded-Distortion that maps $\ell$ to itself while preserving the positive direction as above. Note that this change of coordinates does not change the distortion $\mu_f$ of the affine map. Therefore, the collection $\set{f^*}$ of all affine maps satisfying the assumption of the proposition with general lines is convex.

The consequence of this proposition is that the set of $\mu$-bounded distortion affine transformations that map an epipolar line in one image to an epipolar line in another image is convex, provided that the pair of epipolar lines can be oriented. Consider a pair of epipolar lines $\ell_1$ and $\ell_2$. It can be readily shown that any planar patch in 3D whose front size is visible to both cameras will project to $\ell_1$ and $\ell_2$ with consistent orientation. We note however that for more general scene structures orientation may not always be preserved. Still, many stereo algorithms assume \textit{ordering} (dating back to~\cite{Baker81}). We therefore conclude with the following corollary:
\begin{corollary}\label{prop:basic_epi}
The collection of $\mu$-Bounded-Distortion planar affine transformations that map a directed epipolar line $\ell_1$ to another directed epipolar line $\ell_2$ is convex.
\end{corollary}

\subsubsection{Mappings of triangulations}\label{ss:mapping_triangulations}
\begin{figure}[t]
\includegraphics[width=\columnwidth]{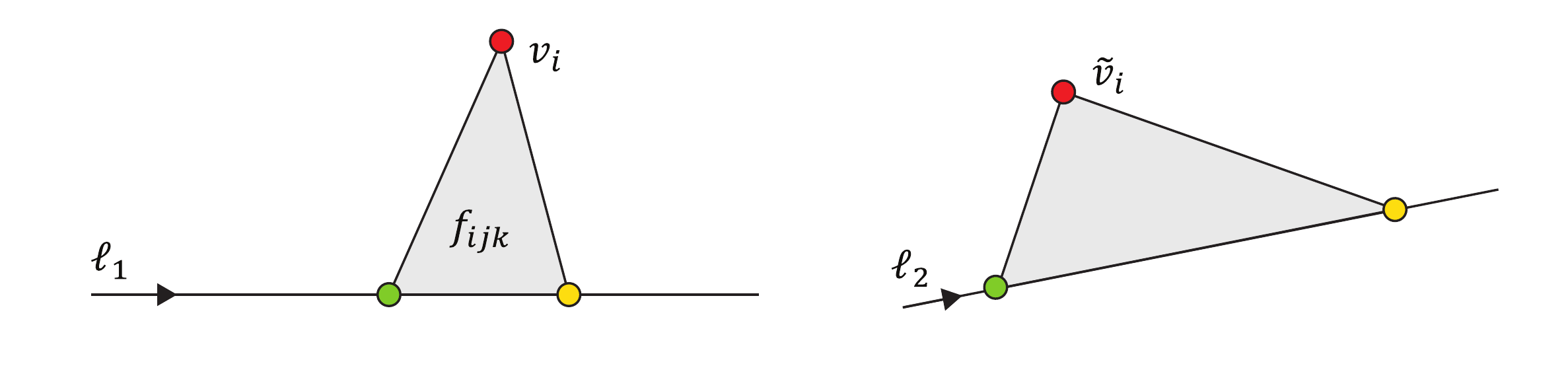}
\caption{Epipolar Bounded-Distortion affine mapping.}\label{fig:epibd}
\end{figure}

We use the results of the previous subsection to formulate our convex mapping space $\bdk$, where each of its members, $\Phi\in\bdk$, is a piecewise linear map whose restriction to a triangle $f_{ijk}\in \F$ is an affine map $\Phi_{ijk}$. Let us denote $$\Phi_{ijk}(\vecx) = B_{ijk}\vecx + C_{ijk}\vecx+\vect_{ijk}.$$
The coefficient of this affine map $B_{ijk}$, $C_{ijk}$, and $\vect_{ijk}$ are all linear functions of the degrees of freedom $\wt{\V}$ (\ie, the mapped vertices) of the mapping $\Phi$ as follows,
\begin{equation}\label{e:linear_relation}
\brac{ B_{ijk}+C_{ijk} \,\, \vert \,\, \vect_{ijk} } \hspace{-0.0cm} =\hspace{-0.0cm} \hspace{-0.1cm}\brac{\begin{array}{ccc}
\hspace{-0.175cm}\wt{\vecv}_i &\hspace{-0.175cm} \wt{\vecv}_j &\hspace{-0.175cm} \wt{\vecv}_k \hspace{-0.175cm}\\
\end{array}} \brac{\begin{array}{ccc}
\hspace{-0.17cm}\vecv_i & \hspace{-0.175cm}\vecv_j &\hspace{-0.175cm} \vecv_k \hspace{-0.175cm}\\
\hspace{-0.17cm}1 & \hspace{-0.175cm}1 &\hspace{-0.175cm} 1 \hspace{-0.175cm} \\
\end{array}}^{-1}
\end{equation}
where here $\vecv_i,\wt{\vecv}_i\in\Real^{2\times 1}$ are viewed as vectors in the plane. Note that the inverted matrix (the rightmost matrix in~\eqref{e:linear_relation}) is constant as it only depends on the source triangulation's vertices $\V$. Therefore, if the triangle $f_{ijk}$ has an edge on an epipolar line $\ell_1$, we can set $\ell_2 = F\ell_1$ with $F$ being the \textit{Fundamental matrix} and combine \eqref{e:linear_relation} with \eqref{e:change_of_coords}, \eqref{e:cone} and \eqref{e:a_and_b} to constrain $\Phi_{ijk}$ to be $\mu$-Bounded Distortion and to respect the epipolar constraint $\ell_1\too\ell_2$. See Figure \ref{fig:epibd} for an illustration. For the third vertex of $f_{ijk}$ (shown in red) we can impose its epipolar constraint by adding the suitable linear equation. Adding these equations for all triangles $t_{ijk}\in \F$ (one SOC and a few linear equality constraints per triangle) results in a convex SOCP realization of the space of PL mappings $\bdk$ with a single distortion parameter $\mu\in(0,1)$.

\subsubsection{Triangulating the source image}
\label{sec:triangulation}

\begin{figure}[t]
\includegraphics[width=\columnwidth]{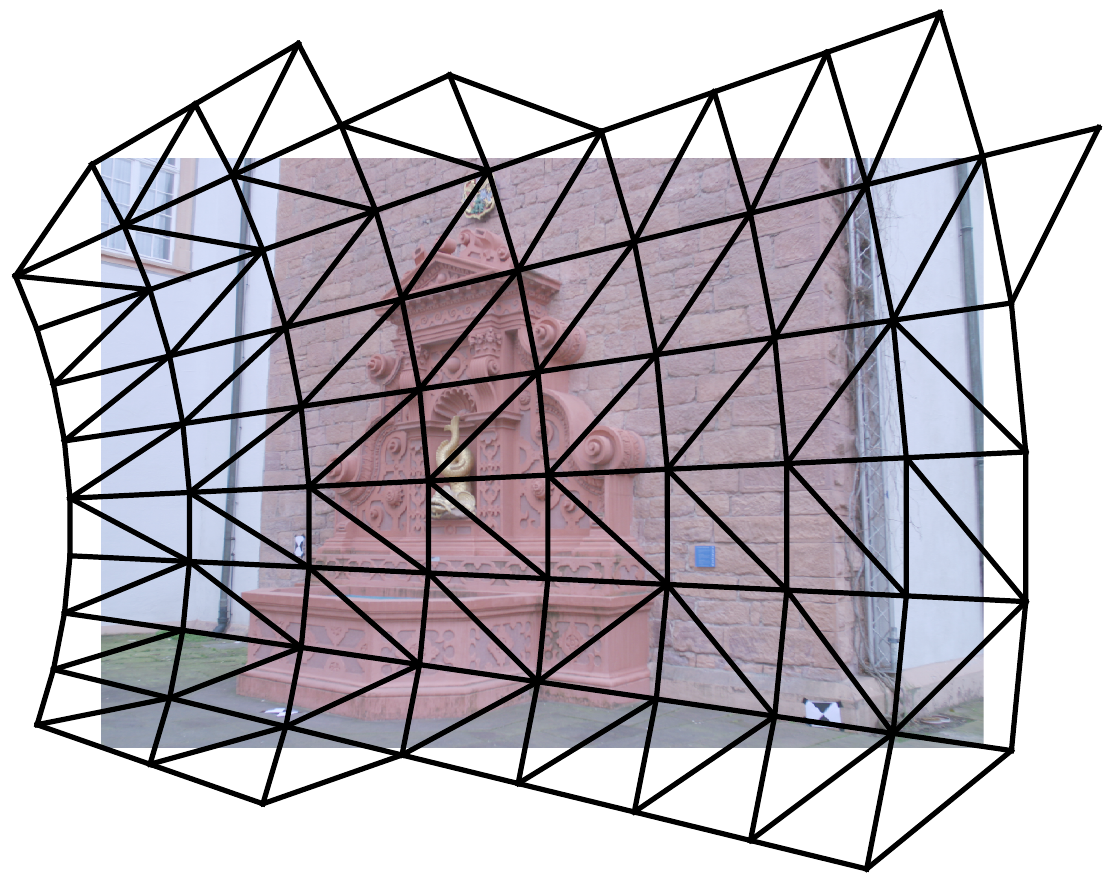}
\caption{Example of an epipolar triangulation of an image. For illustration purposes we show coarse triangles.}\label{fig:triangulation}
\end{figure}

In order to construct $\bdk$ we require a triangulation $\T=(\V,\E,\F)$ with the property that each triangle has an edge on an epipolar line $\ell_1$ of image $I$. We call such a $\T$ an \textit{epipolar triangulation}. We construct such a triangulation by placing an equispaced grid of distance $\eta$ over a polar coordinate frame centered at the epipole (we used $\eta=25$ pixels). For each triangle we enforce its edges to coincide with the appropriate epipolar lines by applying constrained Delaunay triangulation is non-empty. We only keep triangles whose intersection with the image. Figure~\ref{fig:triangulation} depicts an example.
We further determine the orientations of the epipolar lines. This can be done simply by recovering projective camera matrices from the fundamental matrix $F$ and testing the orientation induced, say, by the $Z=const$ plane.

\subsection{Optimization}
To optimize \eqref{e:opt} we first use a simple modification of SIFT~\cite{SIFT} to find candidate pairs of corresponding points $(\vecp_m,\vecq_m)$ that satisfy the epipolar constraint. If the fundamental matrix $F$ is not provided we use standard SIFT and RANSAC to first estimate $F$.

Next, we optimize \eqref{e:opt} using IRLS combined with convex epipolar $\mu$-Bounded Distortion constraints. Assuming a fixed list of pairs $(\vecp_m,\vecq_m)$, we reformulate \eqref{e:opt} as
\begin{subequations} \label{e:irls}
\begin{align} \label{e:irls1}
\min_\Phi & \,\,\,\sum_{m=1}^n  g_{p,\eps}( \norm{\vech_m}) \\ \label{e:irls2}
\mathrm{s.t.} & \,\,\, \vech_m=\Phi(\vecp_m)-\vecq_m \\ \label{e:irls3}
 & \,\,\, \Phi\in \bdk,
\end{align}
\end{subequations}
where $\vech_\ell\in\Real^{2\times 1}$ are auxiliary variables, and the functions $g_{p,\eps}$  will be defined soon. The map $\Phi$ is represented by the images of the vertices of the triangulation $\T$, that is $\set{\wt{\vecv}_i}$. Namely, each vertex $\vecv_i$ is mapped to a new (unknown) location in the second image $\wt{\vecv}_i\in J$, and $\Phi$ is the unique piecewise linear interpolation $\Phi_{ijk}$ over the triangles $f_{ijk}$, as described in Section \ref{ss:mapping_triangulations}. The unknowns in the optimization problem \eqref{e:irls} is therefore the target vertex locations $\set{\wt{\vecv}_i}$.

 The constraint \eqref{e:irls2} is set for every $m$ by finding the triangle $f_{ijk}$ containing $\vecp_m$ and encoding $\vecp_m$ in barycentric coordinates of the corners $\vecv_i,\vecv_j,\vecv_k$ of that triangle, namely $\vecp_\ell=c_{m,i} \vecv_i+c_{m,j}\vecv_j + c_{m,k}\vecv_k$, where the barycentric weights satisfy $c_{m,i},c_{m,j},c_{m,k}\geq 0$ and $c_{m,i}+c_{m,j} + c_{m,k}=1$. \eqref{e:irls2} then becomes
\begin{equation}\label{e:bary}
\vech_m=c_{m,i} \wt{\vecv}_i+c_{m,j}\wt{\vecv}_j + c_{m,k}\wt{\vecv}_k-\vecq_m.
\end{equation}
The EBD constraint \eqref{e:irls3} is set by adding Equations \eqref{e:linear_relation},\eqref{e:change_of_coords},\eqref{e:cone} and \eqref{e:a_and_b} for every triangle $f_{ijk}\in\F$ of the triangulation $\T$. Note that \eqref{e:cone} is a second order cone, and the rest of the equations are linear equalities and inequalities.

Lastly, optimizing the energy \eqref{e:irls1} w.r.t.~$\Phi$ requires to cope with the non-convexity and non-smoothness of the energy \eqref{e:opt1}. The IRLS point of view suggests replacing the zero norm with its approximations \begin{equation}
g_{p,\eps}(r)= \begin{cases}
r^p & r>\eps \\
\frac{p}{2} \eps^{p-2} r^2 + \parr{1-\frac{p}{2}} \eps^p & 0\leq r \leq \eps
\end{cases}
\end{equation}
The $g_{p,\eps}$ functions are smooth ($C^1$) and converge to $r^0$ as $p,\eps\too 0$. For a fixed $p,\eps$, \eqref{e:irls1} is optimized iteratively by replacing $g_{p,\eps}(r)$ with a convex quadratic functional called \emph{majorizer}, $G_{p,\eps}(r,s)$, with the properties that $G_{p,\eps}(s,s)=g_{p,\eps}(s)$, and $G_{p,\eps}(r,s)\geq g_{p,\eps}(r)$, for all $r$. These two properties guarantee that the IRLS monotonically reduces the energy in each iteration. The majorizers $G_{p,\eps}$ are similar to those in~\cite{Bissantz09},
\begin{equation}
G_{p,\eps}(r,s)= \begin{cases}
\frac{p}{2} s^{p-2} r^2 + \parr{1-\frac{p}{2}}s^p & s>\eps \\
\frac{p}{2} \eps^{p-2} r^2 + \parr{1-\frac{p}{2}} \eps^p & 0\leq s \leq \eps
\end{cases}
\end{equation}
Replacing $g_{p,\eps}(\norm{\vech_m})$ in \eqref{e:irls1} with $G_{p,\eps}(\norm{\vech_m},\norm{\vech_{m}'})$, where $\vech_{m}'=\Phi'(\vecp_m)-\vecq_m$, and $\Phi'$ is the map found at the previous iteration, results in the following convex quadratic energy in $\vech_m$ (remember that $\vech'_m$ are constants),
\begin{subequations} \label{e:irls_final}
\begin{align} \label{e:irls_final1}
\min_\Phi & \,\,\,\sum_{m=1}^n w(\|\vech'_m \|) \, \|\vech_m\|^2 \\
\mathrm{s.t.} & \,\,\, \vech_m=\Phi(\vecp_m)-\vecq_m \\
 & \,\,\, \Phi\in \bdk
\end{align}
\end{subequations}
where $w(s) = \max\{s,\eps\}^{p-2}$ is constant at each iteration.
In view of \eqref{e:bary} this implies a convex quadratic energy in the unknowns $\set{\wt{\vecv}_i}$. We iteratively solve this problem, updating $\vech'_j,\Phi'$ in each iteration until convergence. Each iteration is a convex Second Order Cone Program (SOCP) and is solved using MOSEK~\cite{mosek}.

In practice, we fix $p=0.001$ and $\eps$ to be the diameter of image $I$ and solve the above IRLS. Upon convergence, we update $\eps=0.5\eps$ and repeat. We continue this until $\eps=1$ (pixels). This heuristic of starting from a large $\eps$ and decreasing it helps avoiding local minima of the energy \eqref{e:opt1} as the larger the $\eps$ the more convex the problem is; for example, for sufficiently large $\eps$ the global minimum of \eqref{e:irls} lies in the convex (quadratic) part of all terms $g_{p,\eps}$ and can be found by a single SOCP. Our algorithm is summarized in Algorithm~\ref{alg1}.
\begin{algorithm}[t]
   \floatname{algorithm}{Algorithm}
   \begin{algorithmic}[1]


   \Require Two images $I$ and $J$, Fundamental matrix $F$, distortion bound $\mu$, Triangle edge length $\eta$, and a bound on the Sampson Distance $\delta$

   \item[]

   \State // Find putative matches

       $\{(\vecp_m,\vecq_m)\}=\mathrm{EpipolarSIFT}(I,J,F,\delta$)

   \State // Epipolar triangulation of $I$ according to $F$

(Section~\ref{sec:triangulation})

$\T=\mathrm{Delaunay Triangulation}(I,Constraints(F),\eta)$
   \State Compute barycentric coordinates for $\{\vecp_m\}$~\eqref{e:bary}
   \State // Optimization

$p=0.001$, $\epsilon=\mathrm{diameter}(I)$;
\State $\forall m, \, \vech_m'=\vecp_m-\vecq_m$
   \While {$\epsilon \le 1$}
       \While {Not converged}
            \State Solve Eq.~\eqref{e:irls_final} using SOCP solver, obtaining $\Phi$
            \State $\forall m, \, \vech_m'= \Phi(\vecp_m)-\vecq_m$
       \EndWhile
       \State $\epsilon = \epsilon / 2$
   \EndWhile
   \State \Return A subset of matched points $\{(\vecp_{m_i},\vecq_{m_i})\}$ and a map $\Phi$

        \end{algorithmic}

        \caption{ \label{alg1}}

\end{algorithm}

\section{Experiments}

\noindent \textbf{Datasets}. ~
We evaluate our method by applying the optimization algorithm presented in Sec.~\ref{sec:method} to pairs of images from the dataset of~\cite{strecha2008benchmarking}. The dataset contains two multi-view collections of high-resolution images $(2048 \times 3072)$, referred to as ``Herzjesu" and ``Fountain,"  provided with ground truth depth maps. The Herzjesu dataset contains $8$ images and the Fountain dataset contains $11$ images. Therefore, in total there are 83  stereo pairs with varying distances between focal points. We tested each pair twice, seeking a map from the left image to the right one and vice versa, obtaining 166 matching problems.

For evaluation we further process the ground truth depth values to obtain ground truth matches. Specifically, for each dataset we employ ray-casting (z-buffering) to the 3D surface, obtaining ground-truth correspondences at sub-pixel accuracy. We further used ray casting to determine an occlusion mask and excluded those pixels (for the left image) from our evaluation. (These masks of course are not known to the algorithm and used only for evaluation.)

Our optimization algorithm can work in reasonable run-times (roughly 5 minutes) when applied to the high-resolution images. However, in order to compare to state-of-the-art algorithms, which are considerably slower at those resolutions, we use the lower-resolution $(308\times461)$ suggested in~\cite{Daisy_2010,segmentation-aware_2013}. We do not rectify the images or apply any other pre-processing.

\vspace{0.3cm}

\begin{figure}[t]
\begin{center}
\includegraphics[width=7.7cm]{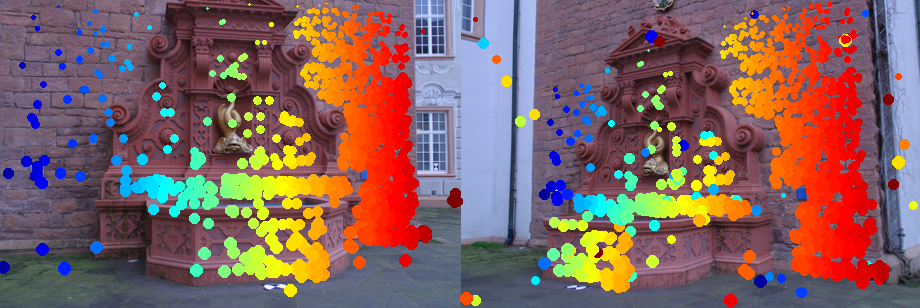}
\end{center}
\caption{Putative matches obtained with the classical SIFT algorithm, which seeks matches over the entire image. The figure shows images 7 and 3 from the Fountain dataset.}\label{fig:classicalSift}
%
\begin{center}
\includegraphics[width=7.7cm]{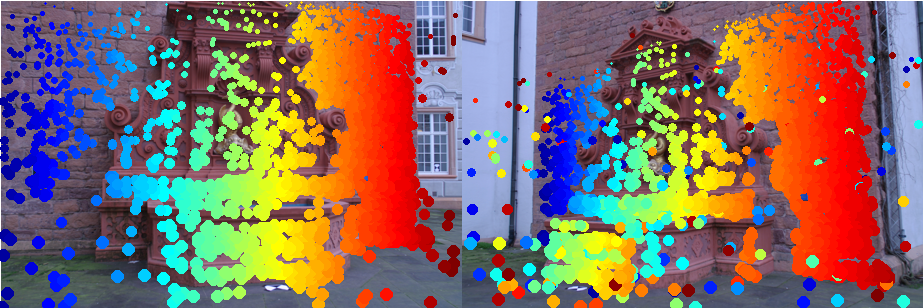}
\end{center}
\caption{Putative matches obtained with Epipolar SIFT. In this case the search for matches is restricted by the Sampson distance  to the immediate surroundings of the corresponding  epipolar line. It is evident that the set of putative matches is richer than that obtained with the matching algorithm, Fig.~\ref{fig:classicalSift}. }\label{fig:epipolarSift}
\end{figure}

\noindent \textbf{Epipolar SIFT}. ~
Our algorithm takes as input pairs of putative correspondences and builds an EBD map that is consistent with as many of the input matches as it can, For the experiments we used SIFT matches (using the VLFeat software package~\cite{VLFeat}). Classical SIFT matching seeks putative matches throughout the {\it entire} image domain. As we assume that epipolar geometry is known (either exactly or approximately), we modify the matching procedure as follows. Given a SIFT descriptor at location $\vecp$ in the left image, we restrict the search for a putative match, $\vecq$, to the area close to the corresponding epipolar line in the right image. This area is determined by limiting the Sampson distance between $\vecp$ and $\vecq$, i.e.
\begin{equation}
\frac{(\vecq^T F \vecp)^2}{(F \vecp)^2_1 + (F \vecp)^2_2 + (F^T \vecq)^2_1 + (F^T \vecq)^2_2} < \delta
\end{equation}
where $F$ is the fundamental matrix,  $\vecp$ and $\vecq$ are written in homogeneous coordinates, and $(F\vecp)_i$ denotes the $i^{th}$ entry of the vector $F\vecp$. We further accept a match $(\vecp,\vecq)$ if its SIFT score is at least twice higher than the score of $(\vecp,\vecq')$ for any $\vecq'$ within Sampson distance $\delta$. We set $\delta$ to 5. Fig.~\ref{fig:classicalSift} shows an example of the putative matches obtained using the classical methodology of SIFT, while Fig.~\ref{fig:epipolarSift} demonstrates the the putative matches obtained with the described methodology, Epipolar SIFT. In these pictures the images are presented side-by-side with the color of the markers corresponding to the value of the $x-$coordinate and the size of the marker corresponds to the value of $y-$coordinate. It is evident that the set of putative matches obtained with Epipolar SIFT is reacher than that obtained with the classical method.

\vspace{0.3cm}

\noindent \textbf{Algorithms for evaluation}. ~
We compare our method to the following algorithms:
\begin{enumerate}

\item \textbf{BD}: Feature matching by bounded-distortion suggested by Lipman et al.~\cite{lipman12}.
This method serves as baseline to our method since it seeks correspondences consistent with a bounded distortion transformation, but does not take epipolar constraints into account.
\item \textbf{Spectral}: The spectral technique of Leordeanu and Hebert~\cite{Leordeanu05}. This method uses graph methods to find point matches by minimizing pairwise energies.

\item \textbf{SiftFlow}: by Liu et al.~\cite{SiftFlow_2011}, which finds dense correspondence by minimizing an MRF energy whose unary term measures the match between SIFT descriptors,

\item \textbf{Homography}: Mapping by looking for the best homography (computed with RANSAC~\cite{Fischler81})

\item \textbf{Stereo}: by Lee et al.~\cite{lee2015robust}, which finds dense correspondence between the images after rectification.
\end{enumerate}
 We note that the algorithms of~\cite{Leordeanu05} and~\cite{SiftFlow_2011} were not designed specifically for wide baseline stereo. For a fair comparison we therefore tested those algorithms in two  settings, first in their original (unrestricted) setting, and secondly in a setting that integrates the knowledge of epipolar geometry into the algorithms. The latter is achieved as follows. For \cite{Leordeanu05} we used a version of the algorithm that allows it to select from a candidate set of matches that were either extracted from the entire image (for the unrestricted setting) or from the epipolar SIFT matches (\ie, the same input given to our algorithm). Furthermore, since this algorithm does not compute a map (it only return a sparse set of matches) we further applied cubic interpolation to extend the matches to the entire image. For \cite{SiftFlow_2011} we modified the code to allow only maps on or close to corresponding epipolar lines (we set the Sampson distance to 2, which gave the best result).
Finally, for homography we used putative matches obtained with the epipolar SIFT and for the stereo algorithm we used ground truth matches to perform the rectification.
\vspace{0.3cm}

\begin{figure}[t]
\begin{center}
\includegraphics[width=7.7cm]{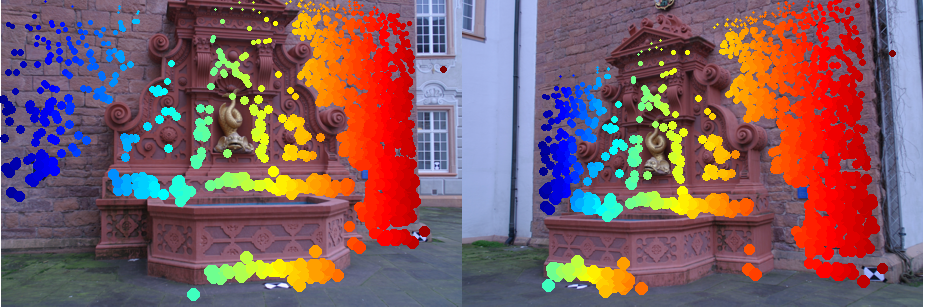}
\end{center}
\caption{Matches $\{(\vecp_m,\vecq_m)\}$ obtained with our EBD solver.}
\label{fig:matches}
\begin{center}
\includegraphics[width=7.7cm]{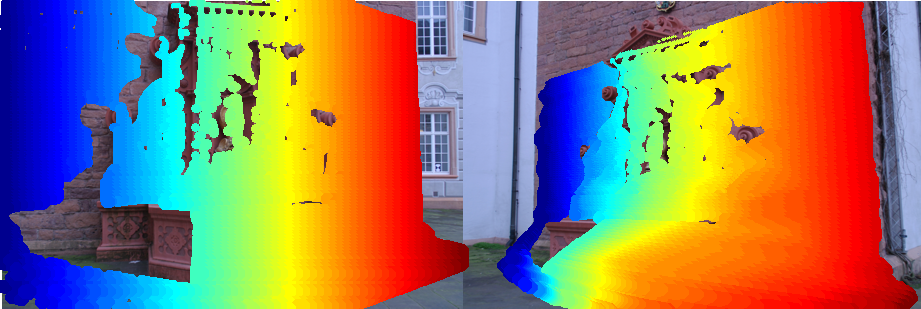}
\end{center}
\caption{The map $\Phi$ obtained with our EBD solver.}
\label{fig:map}
\end{figure}

\begin{figure}[t]
\begin{center}
\includegraphics[width=7.2cm]{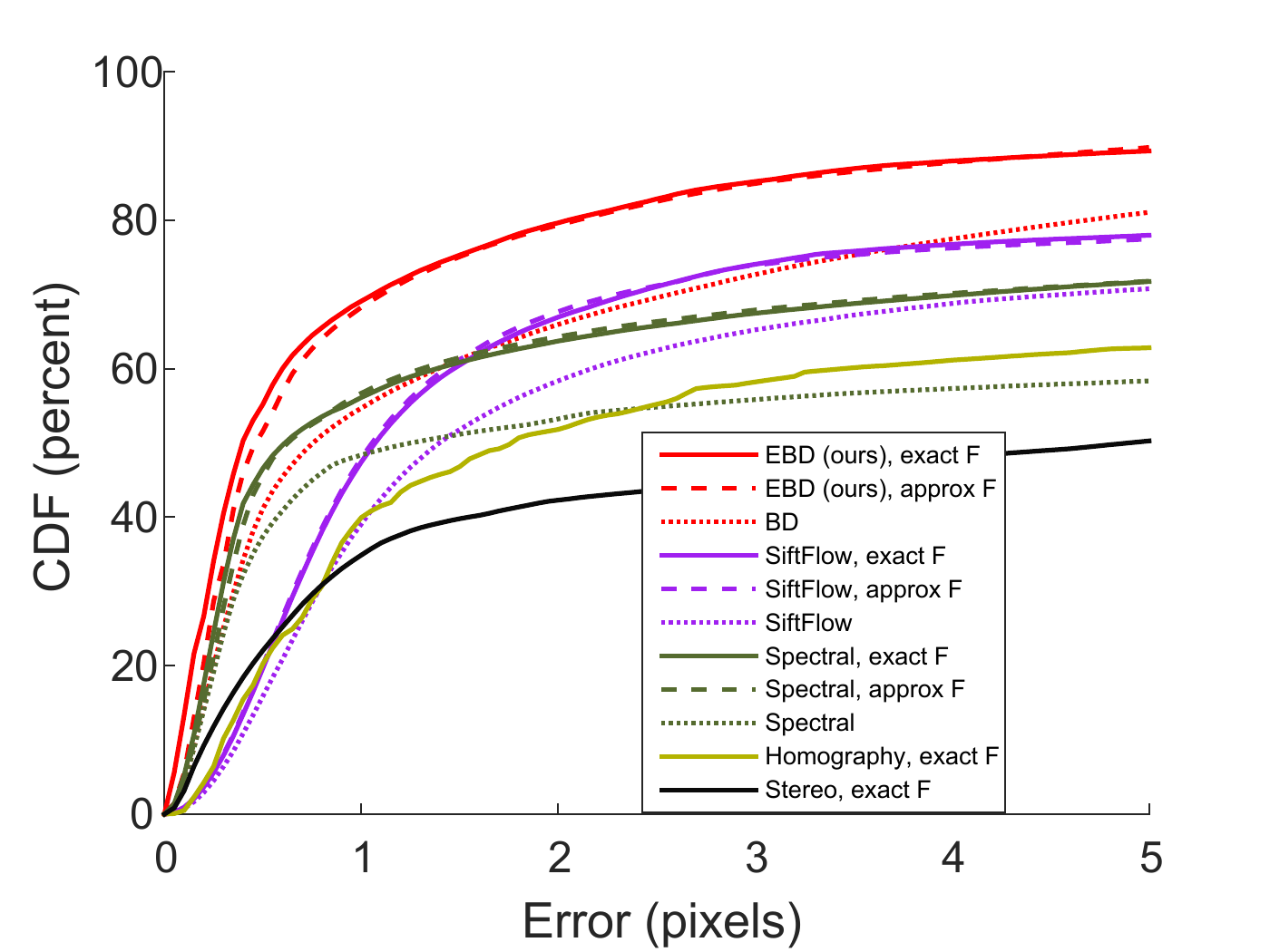}
\end{center}
\caption{The percent of pixels mapped by each method to within an error specified on the horizontal axis from their ground truth target location, for all pairs of images. Median computed for all pairs in the Herzjesu dataset.}
\label{fig:resultsHerzjesu}
%
\begin{center}
\includegraphics[width=7.2cm]{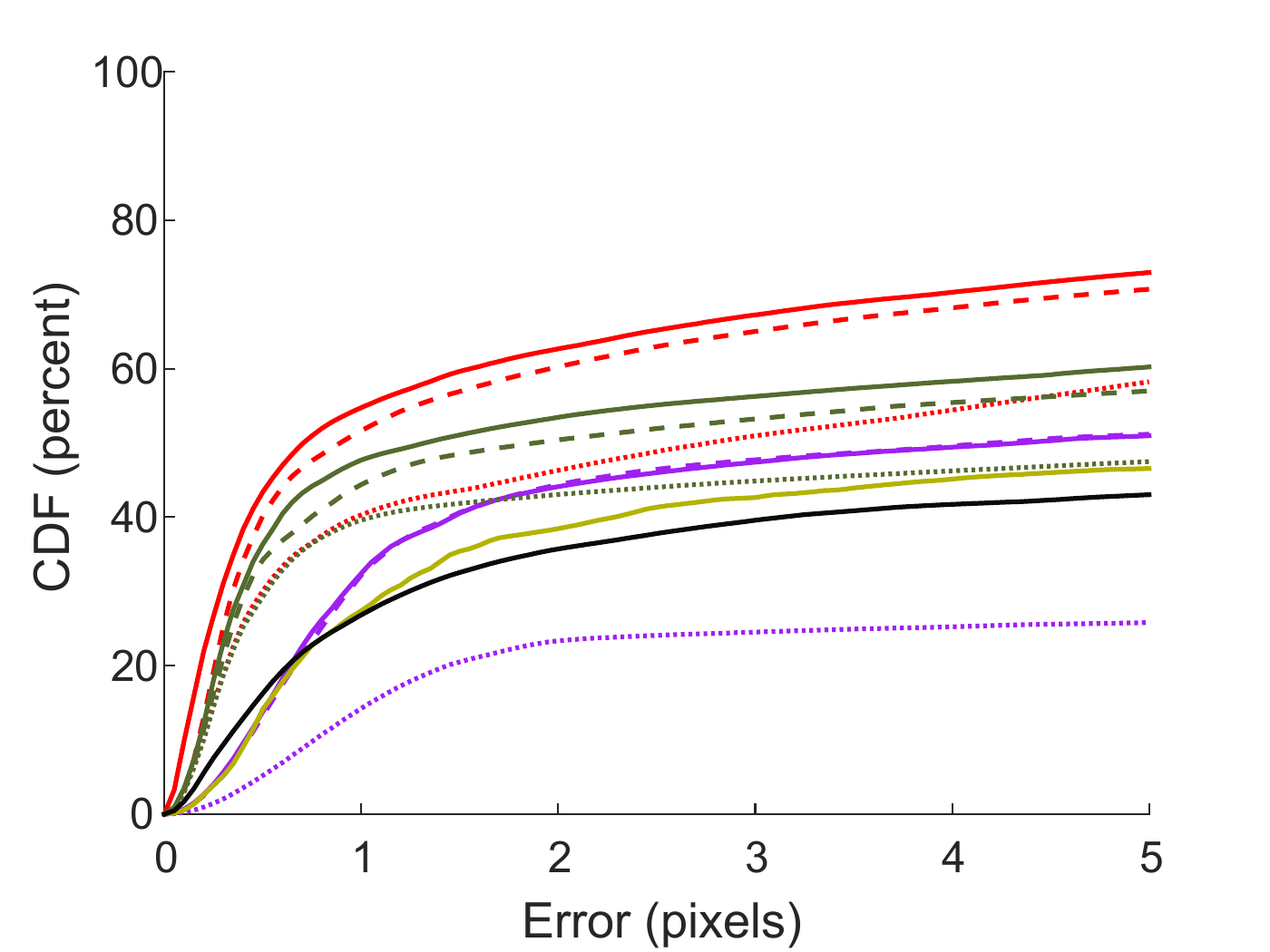}
\end{center}
\caption{The percent of pixels mapped by each method to within an error specified on the horizontal axis from their ground truth target location, for all pairs of images. Median computed for all pairs in the Fountain dataset (legend of Fig.~\ref{fig:resultsHerzjesu} applies here).}
\label{fig:resultsFountain}
\end{figure}

\noindent \textbf{Results}. ~
Figures~\ref{fig:matches} and~\ref{fig:map} show an example for the results obtained with our method. The figures show respectively the set of correspondences $\{\vecp_m,\vecq_m)\}$ and the map $\Phi$ returned by our optimization. To further evaluate the map computed with our algorithm for the entire dataset, we checked for each tested pair of images $I$ and $J$ all pixels in $I$ after masking it with the ground truth occlusion map. For each non-occluded pixel $\vecp$ we measured the Euclidean distance $\|\Phi(\vecp)-\vecq\|,$ where $\vecq$ is the ground truth point corresponding to $\vecp$. We then produced a cumulative histogram depicting the fraction of non-occluded points in $I$ against their displacement error from the ground truth target position. In Figures~\ref{fig:resultsHerzjesu} and~\ref{fig:resultsFountain} we report for each error value the median number of points that achieved this error or less over all pairs of images. Table~\ref{tab:1pixel} further shows the median fraction of non-occluded pixels that were mapped to a 1 pixel accuracy by our map $\Phi$. We show our results both with an exact fundamental matrix (obtained from ground truth) and with an approximated one (computed with RANSAC~\cite{Fischler81} using classical SIFT). Our results are further compared to Spectral~\cite{Leordeanu05}, SiftFlow~\cite{liu2008sift} (both with and without epipolar constraints), to homography estimation and to classical stereo estimation.  (To simplify the table we only include results for the epipolar-enhanced algorithms.) As can be seen from the figures and the table our method outperformed all the tested methods on both datasets with both an exact and an approximate fundamental matrix. We note further that for all algorithms there was no marked difference between the use of exact and approximate fundamental matrix (solid lines vs.\ dashed) and all methods benefited from incorporating epipolar constraints (compare to dotted lines, for non restricted version).

\begin{figure}[t]
\begin{center}
\includegraphics[width=7.2cm]{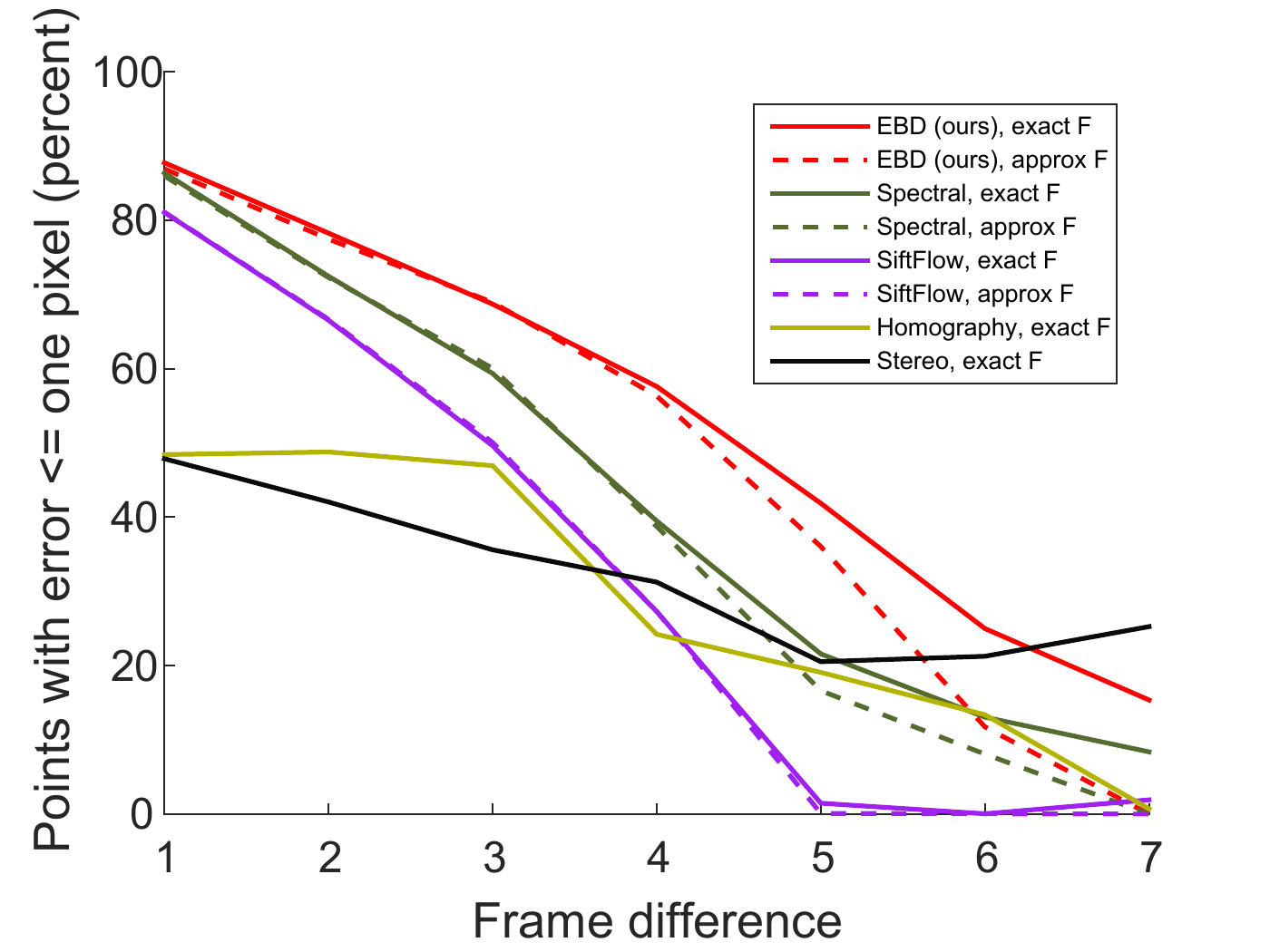}
\end{center}
\caption{Performance as a function of baseline. The graphs shows the percent of pixels mapped by each method to within one pixel from their ground truth target location plotted against frame difference in the sequence for the Herzjesu dataset.}
\label{fig:framesHerzjesu}
%
\begin{center}
\includegraphics[width=7.2cm]{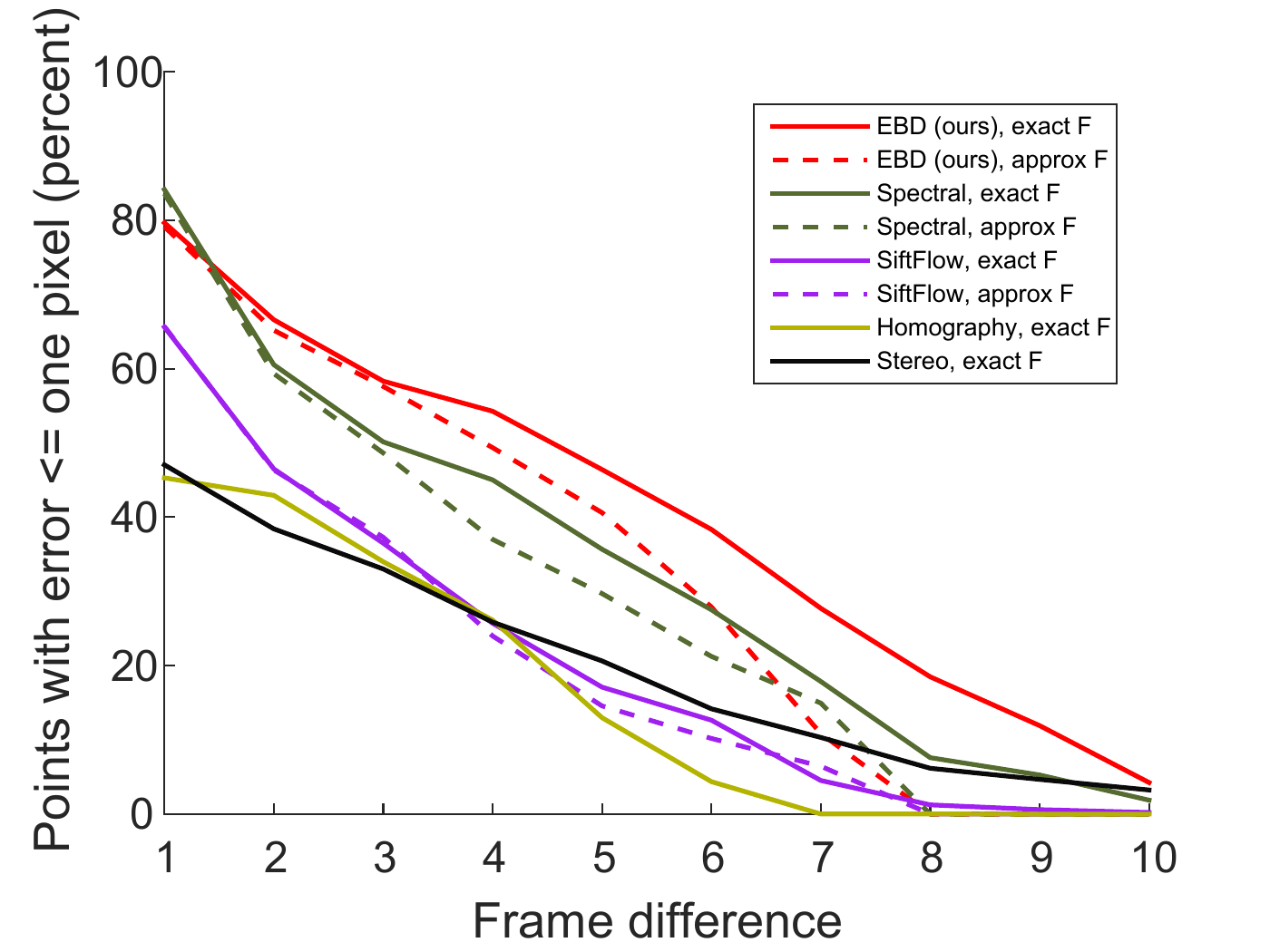}
\end{center}
\caption{Performance as a function of baseline. The graphs shows the percent of pixels mapped by each method to within one pixel from their ground truth target location plotted against frame difference in the sequence for the Fountain dataset.}
\label{fig:framesFountain}
\end{figure}

Figures~\ref{fig:framesHerzjesu} and~\ref{fig:framesFountain} further show a breakdown according to the length of the baseline. For this figure we considered in each of the two datasets all pairs $I_i$ and $I_{i+k}$ for each value $k$ (between 1 and 7 for Herzjesu and between 1 and 10 for Fountain). For each such set of pairs we counted the number of pixels mapped by our computed map $\Phi$ with error $\le 1$ pixel and ploted the median of these numbers. As expected the closer together pairs are, the better our method is. Compared to the other methods our method seem to achieve superior accuracy in almost all conditions.

For a pair of images in this dataset our algorithm runs in 100 seconds on a 3.50 GHz Intel Core i7. This is compared to 400 seconds required for the non-convex BD of~\cite{lipman12}. In general, running the non-convex BD with features restricted to epipolar lines is significantly slower and achieves slightly inferior results.

\begin{table}
\begin{center}
\begin{tabular}{|l|c|c|}
\hline
Algorithm &          Fountain & Herzjesu \\ \hline
\textbf{EBD (ours), exact F} &    \textbf{54.77} &  \textbf{69.11}  \\   \hline
\textbf{EBD (ours), approx F} &   \textbf{51.65} &  \textbf{68.28}  \\   \hline
Spectral, exact F &      47.70 &  56.13  \\   \hline
Spectral, approx F &     44.40 &  56.70  \\   \hline
SiftFlow, exact F &      32.44 &  47.45  \\   \hline
SiftFlow, approx F &     32.19 &  47.97  \\   \hline
Homography, exact F &    27.40 &  39.95  \\   \hline
Stereo, exact F &        26.84 &  34.89  \\   \hline
\end{tabular}
\end{center}
\caption{The percent of pixels mapped by each method to within one pixel from their ground truth target location. Median computed for all pairs of images in the Fountain and Herzjesu datasets. }
\label{tab:1pixel}
\end{table}


{\small
\bibliographystyle{ieee}


\begin{thebibliography}{10}\itemsep=-1pt

\bibitem{mosek}
E.~D. Andersen and K.~D. Andersen.
\newblock {\em {The MOSEK interior point optimization for linear programming:
  an implementation of the homogeneous algorithm}}, pages 197--232.
\newblock Kluwer Academic Publishers, 1999.

\bibitem{Baker81}
H.~H. Baker and T.~Binford.
\newblock Depth from edge and intensity based stereo.
\newblock In {\em Proc. Int. Joint Conf. on Artificial Intelligence}, pages
  631--–636, 1981.

\bibitem{PatchMatch_2009}
C.~Barnes, E.~Shechtman, A.~Finkelstein, and D.~B. Goldman.
\newblock {PatchMatch}: A randomized correspondence algorithm for structural
  image editing.
\newblock {\em ACM Trans. Graph.}, 28(3), Aug. 2009.

\bibitem{FeaturesLineSegments_2005}
H.~Bay, V.~Ferrari, and L.~V. Gool.
\newblock Wide-baseline stereo matching with line segments.
\newblock In {\em CVPR}, 2005.

\bibitem{Berg05}
A.~C. Berg, T.~L. Berg, and J.~Malik.
\newblock Shape matching and object recognition using low distortion
  correspondence.
\newblock In {\em CVPR}, pages 26--33, 2005.

\bibitem{Bissantz09}
N.~Bissantz, L.~Dumbgen, A.~Munk, and B.~Stratmann.
\newblock Convergence analysis of generalized iteratively reweighted least
  squares algorithms on convex function spaces.
\newblock {\em SIAM J. on Optimization}, 19(4):1828--1845, 2009.

\bibitem{brox2009large}
T.~Brox, C.~Bregler, and J.~Malik.
\newblock Large displacement optical flow.
\newblock In {\em CVPR}, pages 41--48, 2009.

\bibitem{Duchenne11}
O.~Duchenne, F.~Bach, I.-S. Kweon, and J.~Ponce.
\newblock A tensor-based algorithm for high-order graph matching.
\newblock {\em PAMI}, 33(12):2383--2395, 2011.

\bibitem{Ferrari_2003}
V.~Ferrari, T.~Tuytelaars, and L.~V. Gool.
\newblock Wide-baseline multiple-view correspondences.
\newblock In {\em CVPR}, 2003.

\bibitem{Fischler81}
M.~Fischler and R.~Bolles.
\newblock Random sample consensus: a paradigm for model fitting with
  applications to image analysis and automated cartography.
\newblock {\em Com. of the ACM}, 24(6):381--395, 1981.

\bibitem{NRDC_2011}
Y.~HaCohen, E.~Shechtman, D.~B. Goldman, and D.~Lischinski.
\newblock Non-rigid dense correspondence with applications for image
  enhancement.
\newblock {\em ACM Trans. Graph.}, 30(4):70:1--70:9, 2011.

\bibitem{hassner2012sifts}
T.~Hassner, V.~Mayzels, and L.~Zelnik-Manor.
\newblock On {SIFT}s and their scales.
\newblock In {\em CVPR}, pages 1522--1528, 2012.

\bibitem{Matas_2004}
M.~U. J.~Matas, O.~Chum and T.~Pajdla.
\newblock Robust wide-baseline stereo from maximally stable extremal regions.
\newblock {\em Image and Vision Computing}, 22(10):761--767, 2004.

\bibitem{kim2013deformable}
J.~Kim, C.~Liu, F.~Sha, and K.~Grauman.
\newblock Deformable spatial pyramid matching for fast dense correspondences.
\newblock In {\em CVPR}, pages 2307--2314, 2013.

\bibitem{lee2015robust}
S.~Lee, J.~H. Lee, J.~Lim, and I.~H. Suh.
\newblock Robust stereo matching using adaptive random walk with restart
  algorithm.
\newblock {\em Image and Vision Computing}, 37:1--11, 2015.

\bibitem{Leordeanu05}
M.~Leordeanu and M.~Hebert.
\newblock A spectral technique for correspondence problems using pairwise
  constraints.
\newblock In {\em ICCV}, volume~2, pages 1482--1489, 2005.

\bibitem{lin2011smoothly}
W.-Y. Lin, S.~Liu, Y.~Matsushita, T.-T. Ng, and L.-F. Cheong.
\newblock Smoothly varying affine stitching.
\newblock In {\em CVPR}, pages 345--352, 2011.

\bibitem{lipman12}
Y.~Lipman.
\newblock Bounded distortion mapping spaces for triangular meshes.
\newblock {\em ACM Trans. Graph.}, 31(4):108:1--108:13, 2012.

\bibitem{lipman14}
Y.~Lipman, S.~Yagev, R.~Poranne, D.~W. Jacobs, and R.~Basri.
\newblock Feature matching with bounded distortion.
\newblock {\em ACM Trans. Graph.}, 33(3):26:1--26:14, 2014.

\bibitem{SiftFlow_2011}
C.~Liu, J.~Yuen, and A.~Torralba.
\newblock Sift flow: dense correspondence across scenes and its applications.
\newblock {\em PAMI}, 33(5):978--994, 2011.

\bibitem{liu2008sift}
C.~Liu, J.~Yuen, A.~Torralba, J.~Sivic, and W.~Freeman.
\newblock {SIFT} flow: dense correspondence across different scenes.
\newblock In {\em ECCV}, pages 28--42, 2008.
\newblock \url{people.csail.mit.edu/celiu/ECCV2008/}.

\bibitem{SIFT}
D.~Lowe.
\newblock Distinctive image features from scale-invariant keypoints.
\newblock {\em IJCV}, 60(2):91--110, 2004.

\bibitem{Zisserman_1998}
P.~Pritchett and A.~Zisserman.
\newblock Wide baseline stereo matching.
\newblock In {\em ICCV}, 1998.

\bibitem{Schaffalitzky_2001}
F.~Schaffalitzky and A.~Zisserman.
\newblock Viewpoint invariant texture matching and wide baseline stereo.
\newblock In {\em ICCV}, 2001.

\bibitem{Strecha_PDE_2003}
C.~Strecha, T.~Tuytelaars, and L.~V. Gool.
\newblock Dense matching of multiple wide-baseline views.
\newblock In {\em ICCV}, 2003.

\bibitem{strecha2008benchmarking}
C.~Strecha, W.~von Hansen, L.~V. Gool, P.~Fua, and U.~Thoennessen.
\newblock On benchmarking camera calibration and multi-view stereo for high
  resolution imagery.
\newblock In {\em CVPR}, pages 1--8, 2008.

\bibitem{Daisy_2010}
E.~Tola, V.~Lepetit, and P.~Fua.
\newblock Daisy: an efficient dense descriptor applied to wide-baseline stereo.
\newblock {\em PAMI}, 32(5):815--830, 2010.

\bibitem{segmentation-aware_2013}
E.~Trulls, I.~Kokkinos, A.~Sanfeliu, and F.~Moreno-Noguer.
\newblock Dense segmentation-aware descriptors.
\newblock In {\em CVPR}, 2013.

\bibitem{Van_Gool_2000}
T.~Tuytelaars and L.~J.~V. Gool.
\newblock Wide baseline stereo matching based on local, affinely invariant
  regions.
\newblock In {\em BMVC}, 2000.

\bibitem{VLFeat}
A.~Vedaldi and B.~Fulkerson.
\newblock Vlfeat vision software.
\newblock \url{www.vlfeat.org}.

\bibitem{Shah_2003}
J.~Xiao and M.~Shah.
\newblock Two-frame wide baseline matching.
\newblock In {\em ICCV}, 2003.

\end{thebibliography}
}

\end{document}